\documentclass[10pt,twocolumn,letterpaper]{article}

\usepackage{wacv}
\usepackage{times}
\usepackage{epsfig}
\usepackage{graphicx}
\usepackage{amsmath}
\usepackage{amssymb}

\usepackage[linesnumbered,ruled]{algorithm2e}
\include{caption}
\usepackage{subfig}
\usepackage{float}
\usepackage{array}
\usepackage{multirow}

\wacvfinalcopy 

\def\wacvPaperID{97} 

\ifwacvfinal\fi
\begin{document}

\title{Going Deeper with Semantics: Video Activity Interpretation using Semantic Contextualization}

\author{Sathyanarayanan Aakur \\
University of South Florida\\
{\tt\small saakur@mail.usf.edu}
\and
Fillipe DM de Souza \\
University of South Florida\\
{\tt\small fillipe@mail.usf.edu}
\and
Sudeep Sarkar \\
University of South Florida\\
{\tt\small sarkar@usf.edu}
}

\maketitle
\ifwacvfinal\thispagestyle{empty}\fi

\begin{abstract}
A deeper understanding of video activities extends beyond recognition of underlying concepts such as actions and objects: constructing deep semantic representations requires reasoning about the semantic relationships among these concepts, often beyond what is directly observed in the data. 
To this end, we propose an energy minimization framework that leverages large-scale commonsense knowledge bases, such as ConceptNet, to provide contextual cues to establish semantic relationships among entities directly hypothesized from video signal. 
We mathematically express this using the language of Grenander's canonical pattern generator theory. 
We show that the use of prior encoded commonsense knowledge alleviate the need for large annotated training datasets and help tackle imbalance in training through prior knowledge. 
Using three different publicly available datasets - Charades, Microsoft Visual Description Corpus and Breakfast Actions datasets, we show that the proposed model can generate video interpretations whose quality is better than those reported by state-of-the-art approaches, which  have substantial training needs. 
Through extensive experiments, we show that the use of commonsense knowledge from ConceptNet allows the proposed approach to handle various challenges such as training data imbalance, weak features, complex semantic relationships and visual scenes.
\end{abstract}


\section{Introduction}
Recent times have seen a tremendous progress in several computer vision tasks such as object and action category recognition in both images and video. 
Efforts have shifted to understanding the visual input beyond simple recognition to generating sentence-based video descriptions~\cite{thomason2014integrating,venugopalan2014translating,yao2015describing,Pan_2016_CVPR,huang2016connectionist} or answering questions about image content~\cite{Wu_2016_CVPR,Tapaswi_2016_CVPR,Kafle_2016_CVPR,Antol_2015_ICCV}. Such approaches point to a need to understand the semantics of the input beyond categorical classification.
The next research frontier is to go beyond what is directly observable in the visual input for uncovering the semantic structure of events. 
Generating semantically coherent interpretations for a given video involves establishing semantic relationships between the atomic elements (or concepts) of the activity, often extending beyond establishing simple pair-wise relationships among concepts detected from the input signal. 
One would have to contextualize the concepts detected, or rather, \emph{hypothesized} from the image and video signal to arrive at coherent semantic interpretations. 

\begin{figure}[h]
\centering
\includegraphics[width=0.95\columnwidth]{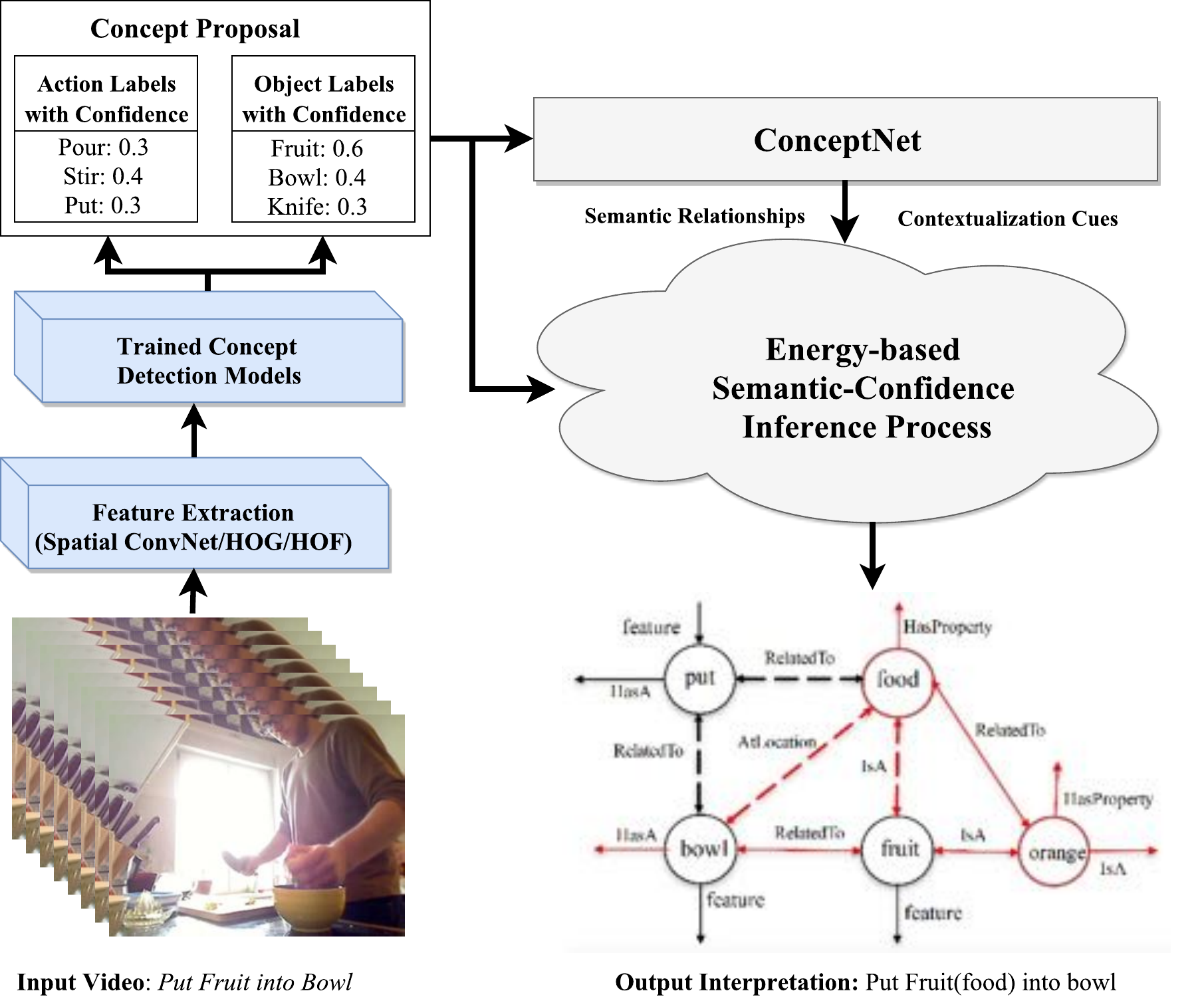}
\caption{\textbf{Overall architecture}: Machine learning-based approaches hypothesize multiple object and action labels. Pattern theory formalism integrates information from ConceptNet to arrive at an interpretation, a semantically coherent structure. Note: Only the modules in blue require explicit training} 
\label{fig:overallArch}
\end{figure}

In this work, we focus on the use of common-sense knowledge bases such as ConceptNet~\cite{liu2004conceptnet,speer2013conceptnet} to capture the underlying semantic structures in video activities. 
We define an \textit{interpretation} as an intermediate representation that forms the basis for generation of more well-formed expressions, such as sentences, or can be the basis for question and answers systems. 
Specifically, an interpretation is a connected structure of basic concepts, such as objects and actions, bound by semantics. 
Similar to scene graphs that are descriptive of static scenes in images~\cite{johnson2015image,xu2017scene,aditya2015images}, interpretations offer a much deeper understanding of the visual scene than labels and can help in constructing meaningful descriptions, answering questions about the scene and retrieve similar events. 
We mathematically express this using the language of Grenander's pattern generator theory~\cite{grenander1996elements}. Concepts are basic generators and the bonds are defined by the semantic relationships between concepts. 
Some concepts in this interpretative structure have direct evidence from video, i.e. grounded concepts, and some are inferred concepts that bind grounded concepts, i.e. contextualization cues (see Section \ref{contextualization}). 
We formulate an inference engine based on energy minimization using an efficient Markov Chain Monte Carlo that uses the ConceptNet in its move proposals to find these structures.

\textbf{Related Work}
Extant approaches have explored the use of context and semantic knowledge in different ways. 
Graphical approaches~\cite{das2013thousand,kuehne2014language}, attempt to explicitly model the semantic relationships that characterize human activities~\cite{lan2012social,kuehne2014language} using a variety of methods such as context-free grammars~\cite{joo2006recognition}, Markov networks~\cite{morariu2011multi}, and AndOr graphs~\cite{wei2013modeling,amer2013monte}. 
These approaches require labeled training data whose sizes increases non-linearly with different semantic combinations of possible actions and objects in the scene. 

Deep learning approaches, employing sequence modeling methods such as RNNs and LSTMs ~\cite{Pan_2016_CVPR,venugopalan2014translating,rohrbach2013translating,yao2015describing,sigurdsson2016asynchronous,shetty2016frame,bin2016bidirectional,dong2016early,guo2016attention,tang2017richer} model the semantic relationships among actions and objects through sequential modeling of such phrases training annotations. 
While effort has been extended to use external text-based resources in addition to the training annotations, they are, arguably, restricted by the quality, quantity, and vocabulary of these annotations. 
This is especially true as the descriptions of visual data are dependent on the sequential semantic relationships encoded in the vocabulary to acquire a strong ontology. 
The acquired ontology of pairwise semantic relationships allows them to handle variations in activity labels within the test and validation sets. 

Our work is a significant departure from extant approaches including those using Grenander's pattern theory~\cite{desouza2016spatially,souzaICPR2016}, which rely on labeled training data to capture semantics about a domain, such as sentences and phrases describing the video segment. 
Unlike them, the applicability of our approach is not restricted to the training domain. 
The use of commonsense knowledge database such as ConceptNet as the source of past knowledge alleviates the need for data annotations. In fact, the only training we require is the one required for detecting underlying concepts such as actions and objects. It allows us to leverage the knowledge gleaned from external sources and hence is not restricted to a particular domain and/or dataset.

\textbf{Contributions:} The contributions of this paper is three-fold: (1) we present a deep semantic reasoning framework for structured representation and interpretation of video activities beyond simple pairwise relationships, 
(2) the use of a global source of knowledge reduces training requirements by negating the need for large amounts of annotations for capturing semantic relationships,  and 
(3) we are, to the best of our knowledge, among the first to introduce the notion of contextualization in activity recognition and show that its introduction improves upon the performance reported by state-of-the-art methods.
\section{Video Interpretation Representation}
Interpreting video content consists of constructing a semantically coherent composition of basic (atomic) elements of knowledge called \texttt{concepts} detected from videos. 
These concepts represent the individual actions and objects that are required to form an interpretation of an activity. 
We use Grenander's pattern theory language~\cite{grenander1996elements} to represent interpretations. 
\subsection{Representing Concepts: Generators}
Following Grenander's notations~\cite{grenander1996elements}, each concept represents a single, atomic component called a \texttt{generator} $g_i \in G_S$ where $G_S$ is the \texttt{generator space}. 
The generator space represents a finite collection of all possible generators that can exist in a given environment. 
In the proposed approach, the collection of more than 3 million concepts in ConceptNet (Section~\ref{ConceptNetSection}), is the generator space.

The generator space ($G_S$) consists of three disjoint subsets that represent three kinds of generators - feature generators ($F$), grounded concept generators ($G$) and ungrounded context generators ($C$). 
Feature generators $(g_{f_1}, g_{f_2},\ldots, g_{f_q} \in F)$ correspond to the features extracted from videos and are used to infer the presence of the basic concepts (actions and objects) called \texttt{grounded concept generators} ($\underline{g}_1, \underline{g}_2, \ldots, \underline{g}_k \in G$). 
Individual units of information that represent the background knowledge of these grounded concept generators are called \texttt{ungrounded context generators} ($\bar{g}_1, \bar{g}_2, \ldots,\bar{g}_q \in C$). 
The term \textit{grounding} is used to distinguish between generators with direct evidence in the video data and the inferred knowledge elements of these concepts.
\subsection{Connecting Generators: Bonds}
\begin{figure}[h]
\centering
\begin{tabular}{c}
\includegraphics[width=1.0\columnwidth]{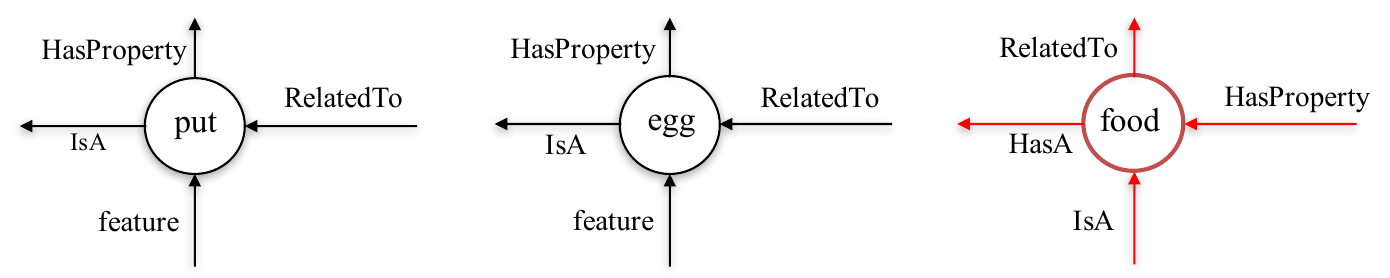} \\ 
(a) \\
\includegraphics[width=1.0\columnwidth]{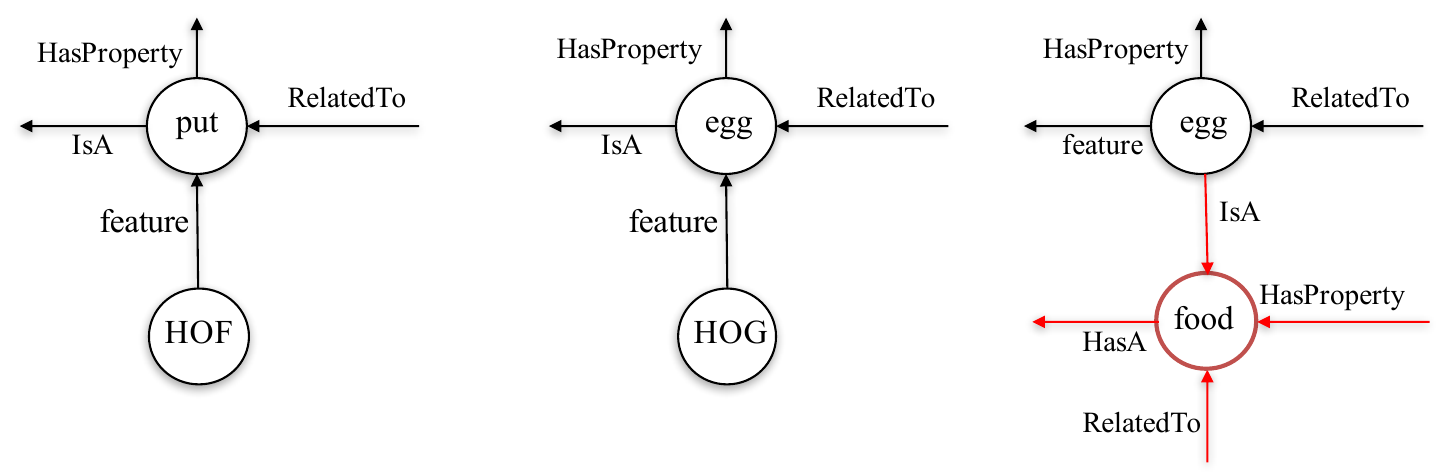}  \\ 
(b) \\
\end{tabular}
\caption{An illustration of generators and their bond structures. 
(a) gives the structure of individual generators. Black generators represent grounded generators and red represents ungrounded generators.
(b) represents bonded pairs of generators.
}
\label{genDescription}
\end{figure}

Each generator $g_i$ has a fixed number of bonds called the \texttt{arity} of a generator ($w(g_i) \forall g_i \in G_S$). 
Bonds are differentiated at a structural level by the direction of information flow that they represent - \textit{in-bonds} and \textit{out-bonds} as can be seen from Figure \ref{genDescription} (a) where the bonds representing \textit{RelatedTo} and \textit{feature} represent in-bonds and \textit{HasProperty} and \textit{IsA} represent out-bonds for the generator \textit{egg}.
Each bond is identified by a unique coordinate and bond value taken from a set $B$ such that the \emph{$j^{th}$} bond of a generator $g_i \in G_S$ is denoted as $\beta_{dir}^{j}(g_i)$, where $dir$ denotes the direction of the bond. 

\textbf{Types of Bonds} There exist two types of bonds - \texttt{semantic bonds} and \texttt{support bonds}. 
\textit{Semantic bonds} are a representation of the semantic relationship between two concept generators. These bonds represent the semantic assertions from ConceptNet. 
The direction of \textit{semantic bonds} signify the semantics of a concept and the type of relationship shared with its bonded generator. 
For example, in Figure \ref{genDescription}(b), the bond \textit{IsA} between concepts \textit{egg} and \textit{food} is a symbolic representation of the semantic assertion that \textit{Egg is a (type of) Food}, signified by the direction of the bond. 
The bonds highlighted in red indicate the presence of ungrounded context generators, representing the presence of contextual knowledge. 
Semantic bonds are quantified using the strength of the semantic relationships between generators through the bond energy function:
\begin{equation}
a_{sem}(\beta^{\prime}(g_i),\beta^{\prime\prime}(g_j)) = \tanh(\phi(g_i,g_j)). 
\label{SemEnergy}
\end{equation}
where $\phi(.)$ is derived from the strength of the assertion in ConceptNet between concepts $g_i$ and $g_j$ through their respective bonds $\beta^{\prime}$ and $\beta^{\prime\prime}$.  
The $\tanh$ function normalizes the output from $\phi(.)$ to range from -1 to 1. 
This is important to note as there can exist negative assertions between two concepts that are not compatible and hence reduces the search space by avoiding interpretations with contrasting semantic assertions.

\textit{Support bonds} connect (grounded) concept generators to feature generators that represent direct image evidence. These bonds are used to preserve the provenance of the concepts with direct data-based evidence. 
Support bonds are quantified through the bond energy function:
\begin{equation}
a_{sup}(\beta^{\prime}(g_i),\beta^{\prime\prime}(g_j)) = \tanh(f(g_i,g_j)). 
\label{SupEnergy}
\end{equation}
where $f(.)$ is derived from the confidence scores of classification models between feature generators $g_i$ and the respective concept generator $g_j$ through their respective bonds $\beta^{\prime}$ and $\beta^{\prime\prime}$. 

A bond is said to be \texttt{open} if it is not connected to another generator through a bond; i.e. an out-bond of a generator $g_i$ is connected to a generator $g_j$ through one of its in-bonds or vice versa. 
For example, take the first case from Figure \ref{genDescription} (b) representing the bonded generator pair \{\textit{egg} and \textit{HOG}\}. 
The bonds representing \textit{HasProperty}, \textit{IsA} and \textit{RelatedTo} are considered to be \textit{open}, whereas the bond representing \textit{feature} represents a \textit{closed} bond. 
\subsection{Interpretations: Configurations of Generators}

\begin{figure*}[h]
\centering
\begin{tabular}{c c}
\includegraphics[width=0.75\columnwidth]{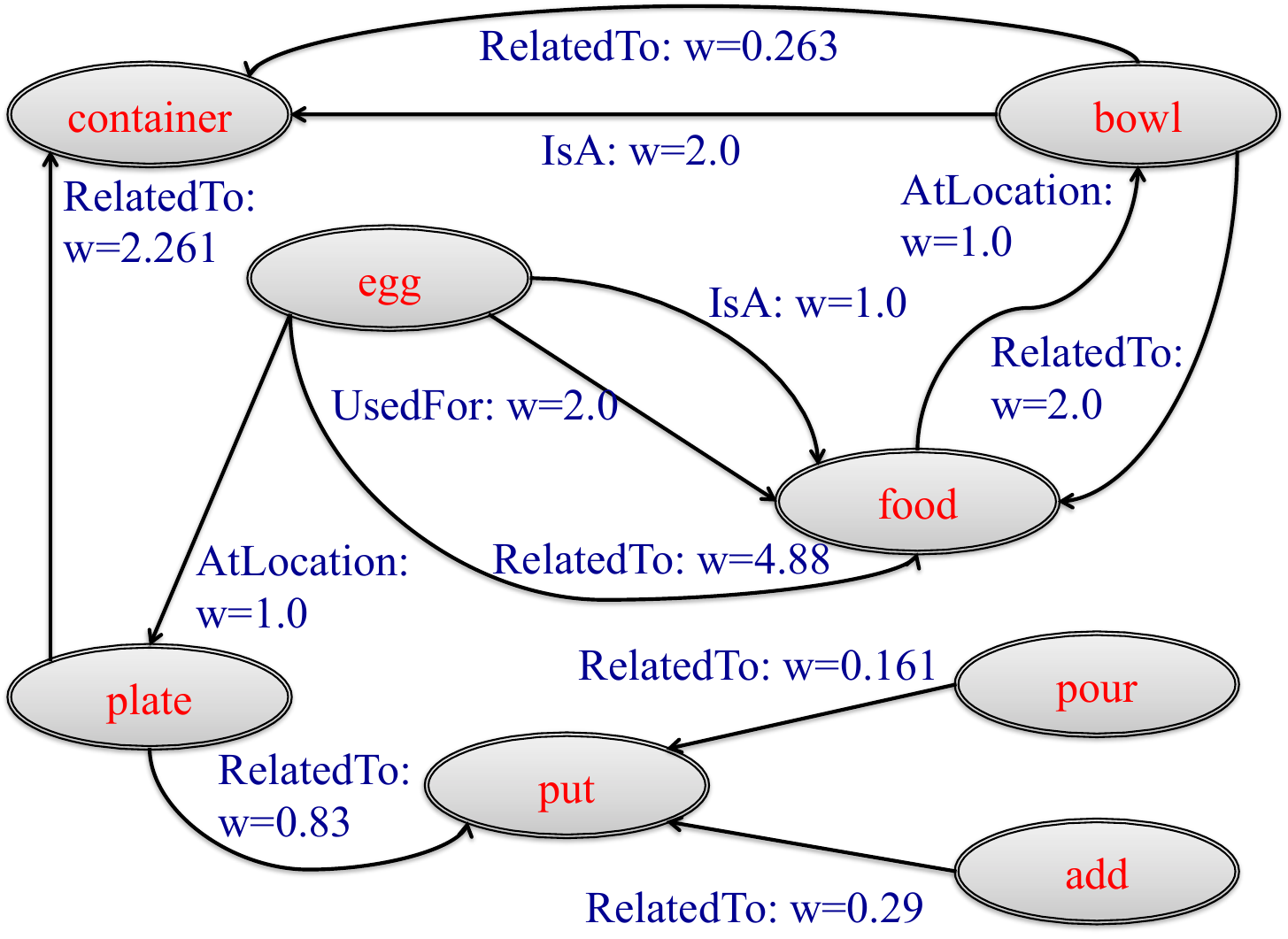}  
& 
\includegraphics[width=0.95\columnwidth]{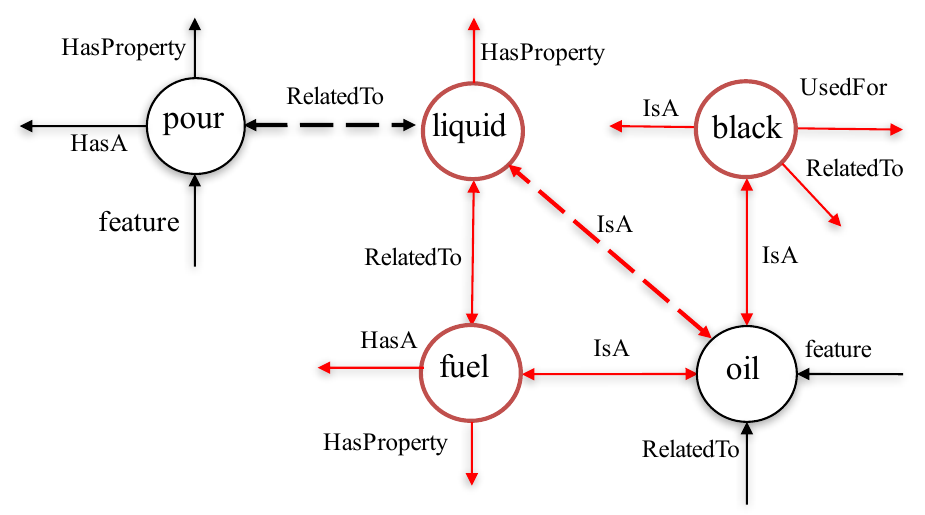} \\
(a) & (b)\\
\end{tabular}
\caption{(a) shows a \emph{tiny} snippet of ConceptNet to show how semantic relationships between concepts are expressed. (b) is a representation of an interpretation using pattern theory. Grounded concepts are represented in black while ungrounded concepts are in red. 
The dashed links represent the optimal semantic relationship between two grounded concepts
}
\label{fig:conceptNetExample}
\label{fig:conceptNetContext}
\end{figure*}

Generators can be combined together through their local bond structures to form structures called  \textit{configurations} $c$, which represent semantic interpretations of video activities (see Figure~\ref{fig:conceptNetContext}(b). 
Each configuration has an underlying graph topology, specified by a connector graph $\sigma$. 
The set of all feasible connector graphs $\sigma$ is denoted by $\Sigma$, also known as the connection type. 
Formally, a configuration $c$ is a connector graph $\sigma$ whose sites $1, 2, \ldots, n$ are populated by a collection of generators $g_1, g_2, \ldots, g_n$ expressed as $c = \sigma (g_1, g_2, \ldots, g_i); g_i \in G_{S}$.
The collection of generators $g_1, g_2, \ldots, g_i$ represents the semantic content of a given configuration c. For example, the collection of generators from the configuration in Figure \ref{fig:conceptNetContext}(b) gives rise to the semantic content ``\textit{pour oil (liquid) (fuel) (black)}''. 

\subsection{From Interpretations to Captions and Labels}\label{genCaption}
Interpretations can be used as a source of knowledge to generate more conventional, text-based expressions such as captions and labels and allows us to perform multiple tasks such as generating descriptions and activity recognition, respectively. 
We generate captions using the template based framework proposed by Thomason et al ~\cite{thomason2014integrating}, which helps transfer the interpretations into a sentence-based expression. 
The template used to generate captions is of the form: \textit{“Determiner (A,The) - Subject - Verb (Present, Present Continuous) - Preposition - Determiner (A,The) - Object”.} 
Each generated sentence is ranked using the BerkeleyLM language model~\cite{pauls2011faster} trained on the GoogleNgram corpus. 
The sentence with the highest average probability is chosen as the final output. 
In our experiments, we find that ungrounded concept generators, while allowing for better expression of semantics, do not add much in terms of evaluation in video caption metrics such as BLEU and METEOR. 
Interpretations can also be converted into labels for activity recognition by a similar template-based format which is given by ``\textit{verb object}''. There is no need for normalizing using a language model as there are no other concepts such as prepositions and determiners in the activity recognition task.
\vspace{-0.20cm}
\section{ConceptNet: Semantic Knowledge Source}\label{ConceptNetSection}
We propose the use of a commonsense knowledge base as a source for learning and establishing semantic relationships among concepts. 
ConceptNet, proposed by Liu and Singh~\cite{liu2004conceptnet} and expanded to ConceptNet5~\cite{speer2013conceptnet}, is one such knowledge base that maps concepts and their semantic relationships in a traversable semantic network structure. With over 3 million concepts, the ConceptNet framework serves as a source of cross-domain semantics supporting commonsense knowledge as expressed by humans in natural language. 
Technically, it encodes knowledge in a hypergraph, with the nodes and edges representing concepts semantic assertions respectively.

There are more than 25 relations (assertions) by which the different nodes are connected, with each of these relations contributing to the semantic relationship between the two concepts such as \texttt{HasProperty}, \texttt{IsA}, and \texttt{RelatedTo} to name a few. 
Each relation has a weight that determines the degree of validity of the assertion given the sources and hence provides a quantifiable measure of the semantic relation between concepts. 
Positive values indicate positive assertions and negative values indicate the opposite. 
Figure~\ref{fig:conceptNetExample}(a) illustrates these ideas; for example, the edge between nodes \texttt{egg} and \texttt{plate} represents an assertion with the relation \texttt{AtLocation} to indicate that eggs can be placed or found in plates. 
We also use these relations to link concepts detected in videos via contextual cues to form semantically coherent interpretations. 

\subsection{Contextualization using ConceptNet}\label{contextualization}
As defined by Gumperz~\cite{gumperz1992contextualization}, primarily for linguistics, contextualization refers to the use of knowledge acquired from past experience to retrieve \textit{presuppositions} required to maintain involvement in the current task. In the context of video interpretation, it refers to the integration of past knowledge to aid in interpreting activities in videos. More specifically, ``\textit{concept}'' refers to actions and objects that constitute an activity; ``\textit{presuppositions}'' refers to the background knowledge of concepts, their properties and semantics. Note that the goal is to generate \textit{interpretations} of a given activity rather than just simple recognition.

Contextualization cues can be of two types - semantic knowledge of basic concepts and relational connection of the current activity to similar ones in the recent past.  The former, explored further in this paper, provides background knowledge about detected objects and actions to aid the interpretation of the video content. It leads to an understanding of \emph{why} entities interact with each other. The latter, on the other hand, allows us to anticipate the entities that may be present in the current context based on relational and temporal cues from experience in the recent past. Combining both types of knowledge derived from contextualization allows us to propose models that are able to express their inferred knowledge through meaningful use of language that leads to better human-machine interaction. 

Two concepts that do have a direct relationship can be correlated using contextualization cues. 
Semantic assertions, particularly \texttt{IsA} and \texttt{HasProperty}, are cues that connect concepts with others which are representative of their contextual properties.
For example in Figure~\ref{fig:conceptNetExample}(a), there is no a direct semantic relationship between the concepts \texttt{egg} and \texttt{put}, but the contextualization cue \texttt{plate} connects them and provides a semantic context, i.e. eggs can be put on plate. 
Contextualization cues allow us to capture the semantic relationships among concepts which we could not have obtained if we rely only on semantic assertions that directly connect concepts. 
Formally, let concepts be represented by $g_i$ for $i=1, \ldots, N$ and let ${_{g_i}R_{g_j}}$ represent relations between two concepts, then contextualization cue, $g_k$, satisfies the following expression $  \mbox{not} \left( {_{g_i}R_{g_j}} \right) \wedge  {_{g_i}R_{g_k}} \wedge {_{g_k}R_{g_j}}$. 

\section{Inference: Constructing Interpretations}
Searching for the best semantic description of a video involves maximizing the probability of a configuration and hence minimizing an energy function $E(c)$. 
The solution space spanned by the generator space is very large as both the number of generators and structures can be variable. 
For example, the combination of a single connector graph $\sigma$ and a generator space $G_S$ give rise to a space of feasible configurations $C(\sigma)$. 
While the structure of the configurations $c \in C(\sigma)$ is identical, their semantic content is varied due to the different assignments of generators to the sites of a connector graph $\sigma$.  

\textbf{Configuration Probability} The probability of a particular configuration $c$ is determined by its energy as given by the relation
\begin{equation}
P(c) \propto  e^{-E(c)}
\label{probConfigEqn}
\end{equation}
where $E(c)$ represents the total energy of the configuration $c$. 
The energy \emph{\(E(c)\)} of a configuration $c$ is the sum of the bond energies (Equations \ref{SemEnergy} and \ref{SupEnergy}) formed by the bond connections between generators in the configuration. 

\begin{equation}
\begin{split}
E(c) &= -\sum_{ (\beta^{\prime},\beta^{\prime\prime})\in c}{a_{sup}(\beta^{\prime}(g_i),\beta^{\prime\prime}(g_j))}\quad + \\
&{a_{sem}(\beta^{\prime}(g_i),\beta^{\prime\prime}(g_j))} + Q(c)
\end{split}
\label{energy}
\end{equation}
where $Q$ is the cost factor associated with using ungrounded context generators as seen from Equation \ref{CostFunction}; 

The energy of a configuration is determined by the energy of all the individual bonds present in the configuration. The first term in Equation \ref{energy} is a reflection of the confidence of the underlying machine learning models, while the second term is a reflection of the semantic coherence of the configuration. 
The third term, the cost factor Q of a given configuration $c$ is given by
\begin{equation}
Q(c) = k \sum_{\bar{g}_i \in G^{\prime}} \sum_{\beta_{out}^{j} \in \bar{g}_i} [D(\beta_{out}^{j}(\bar{g}_i))]
\label{CostFunction}.
\end{equation}
where $G^{\prime}$ is a collection of ungrounded contextual generators present in the configuration $c$, $\beta_{out}$ represents each \textit{out-bond} of each generator $g_i$ and D(.) returns is function that true of the given bond is open. 
$k$ is an arbitrary constant that scales the extent of the detrimental effect that the ungrounded context generators have on the quality of the interpretation.
The cost factor $Q(c)$ restricts the inference process from constructing configurations with degenerate cases such as those composed of unconnected or isolated generators that do not have any closed bonds and as such do not connect to other generators in the configuration. 

A feasible optimization solution for such an exponentially large search space is to use a sampling strategy. We follow the work in~\cite{desouza2016spatially} and employ a Markov Chain Monte Carlo (MCMC) based simulated annealing process. 
The MCMC based simulation method requires two types of proposal functions - global and local proposal functions.
\begin{algorithm}
    \SetKwInOut{Input}{Input}
    \SetKwInOut{Output}{Output}

    localProposal $(c,m)$\;
    Randomly select $g_k \in c$ \\
    Form a set $G^{\prime}$ of $m$ generators $g_i$ such that $g_i \in G_S$ and $g_j \neq g_i$ \\
    Form a set $C^{\prime}$ of generators $\{\bar{g}_j\}$ such that $\bar{g}_j \in C$ and ${_{g_i}R_{\bar{g}_j}}$ exists $\forall g_i \in G^{\prime}$ \\
    Remove $g_j$ from $c$ \\
    Remove $\{\bar{g}_j\} \in C^{\prime}$ from $c$ such that there exists ${_{\bar{g}_i}R_{g_j}}$ \\
    $c^{\prime} \leftarrow c$\\
    Select generator $g_i$ that minimizes $E(\sigma(c^{\prime},g_i))$ \\
    Add $g_i$ to $c^{\prime} $ \\
    Add $\{\bar{g}_k\} \in C^{\prime}$ to $c$ such that there exists ${_{g_i}R_{\bar{g}_k}}$ \\
    $c^{\prime\prime} \leftarrow c^{\prime}$ \\
    return $c^{\prime\prime}$
  \caption{Local Proposal Function}
  \label{localProp}
\end{algorithm}
A connector graph $\sigma$ is given by a global proposal function which makes structural changes to the configuration that are reflected as jumps from a subspace to another. 
To this end, we follow the global proposal from~\cite{desouza2016spatially}. 
A swapping transformation is applied to switch the generators within a configuration to change of semantic content of a given configuration $c$. 
Algorithm \ref{localProp} shows the local proposal function which induces the swapping transformation. 
This results in a new configuration $c^{\prime}$, thus constituting a move in the configuration space $C(\sigma)$. 

Initially, the global proposal function introduces a set of grounded concept generators derived from machine learning classifiers.  
Then, a set of ungrounded context generators, representing the contextualization cues, are populated for each grounded concept within the initial configuration. 
Bonds are established between compatible generators when each generator is added to the configuration. 
Each jump gives rise to a configuration whose semantic content represents a possible interpretation for the given video. 
Interpretations with the least energy are considered to have a higher probability of possessing more semantic coherence. 
We consider the best 10 of the interpretations found over the whole
search trace as the final interpretations of the video.
\section{Implementation Details} \label{grounded}
The detection or rather \emph{hypothesis} of grounded concept generators (possible object and action labels) from the video input represents the only training process. 
We begin with putative object and action labels for each region in the image or video, which are our concepts with direct evidence from data. 
To allow for uncertainty in labeling due to variations in appearance such as occlusion as well variations in temporal scales, we consider top-k ($k=5$ in our experiments) possible labels for each instance. 

To demonstrate the flexibility of the proposed framework, we consider both handcrafted and deep features to detect and label the basic concepts such as actions and objects. 
We experiment with different strategies for feature-level representation of actions and objects. 
First, we follow the work in ~\cite{venugopalan2014translating} and use mean-pooled values extracted from $fc_{7}$ layer for each frame from a CNN model pre-trained on a subset of the ImageNet dataset ~\cite{russakovsky2015imagenet}. 
Second, we leverage more sophisticated features from optical flow and RGB frames in a two-stream architecture as proposed in ~\cite{simonyan2014two} for the Charades dataset. 

Handcrafted features consist of Histogram of Optical Flow (HOF) \cite{chaudhry2009histograms} for generating action labels and Histogram of Oriented Gradients (HOG) \cite{dalal2005histograms} for object labels. 
HOF features were extracted by computing dense optic flow frames from three temporally sequential segments and a histogram of optic flow is constructed for each.
The composite feature for action recognition is composed by concatenating the HOFs. 
Labels were generated using linear support vector machines.
\section{Experimental Evaluation}
We begin with discussion on the three publicly available datasets that we use, followed by presentation of qualitative and quantitative results on them. Performance is quantified using measures that facilitate the comparison with other approaches, such as precision, METEOR and BLEU score~\cite{papineni2002bleu}. It is to be noted that these measures are only a partial reflection of the capability of the interpretations generated by the proposed pattern theory approach. The graphical structure is richer in the conveyed semantics compared to the sentences.

\textbf{Ablation studies}: 
For evaluating the impact of contextualization on the inference process, we use simple, pairwise semantic relationships given by the ConceptNet Similarity edge weights ("PT+weights"), that does not use contextualization. The "PT+weights" approach uses only pairwise relationships that exist among concepts in ConceptNet. If there does not exist a valid semantic relationship between two concepts, then the corresponding semantic assertion is set to zero.
\subsection{Datasets}
The Charades dataset \cite{sigurdsson2016hollywood} is a challenging benchmark containing 9,848 videos across 157 action classes with 66,500 annotated activities. Complex visual data in the form of simultaneous activities and complex semantics offer a considerable challenge. We use the same splits for training and testing from \cite{sigurdsson2016hollywood,sigurdsson2016asynchronous} and evaluate video classification using the evaluation criteria from \cite{sigurdsson2016hollywood} for fair comparison.

The Microsoft Video Description Corpus (MSVD) is a publicly available dataset that contains 1,970 videos taken from YouTube. On an average, there 40 English descriptions available per video. We follow the split proposed in prior works~\cite{thomason2014integrating,venugopalan2014translating,yao2015describing,Pan_2016_CVPR}, and use 1,200 videos for training, 100 for validation and 670 for testing. We evaluate the quality of the generated captions using METEOR and BLEU scores. 

The Breakfast Actions dataset consists of more than 1000 recipe videos, consisting of different scenarios with a combination of 10 recipes, 52 subjects and differing viewpoints, which provides varying qualities of videos. 
The units of interpretation are temporal video segments of these videos, given by the video annotation provided along with the dataset. 

\begin{figure*}[!ht]
\centering
\begin{tabular}{c c c c}
Ground truth  &  Output Interpretation & Ground truth  &  Output Interpretation  \\
\includegraphics[width=0.30\columnwidth]{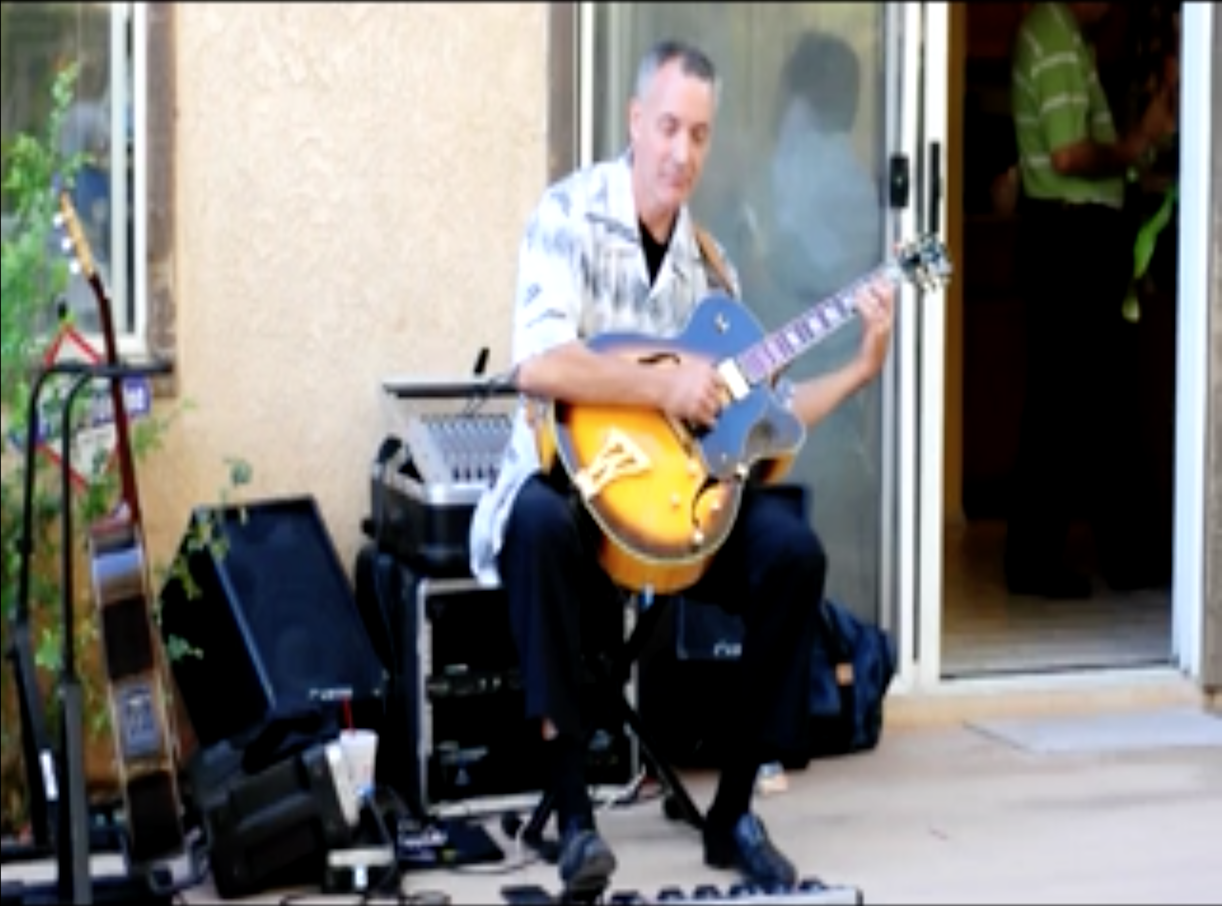} & 
\includegraphics[width=0.55\columnwidth]{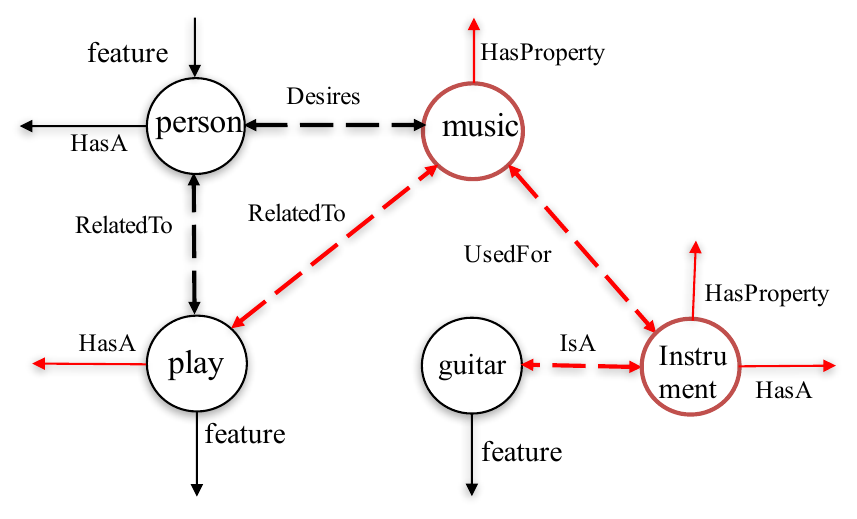} &

\includegraphics[width=0.30\columnwidth]{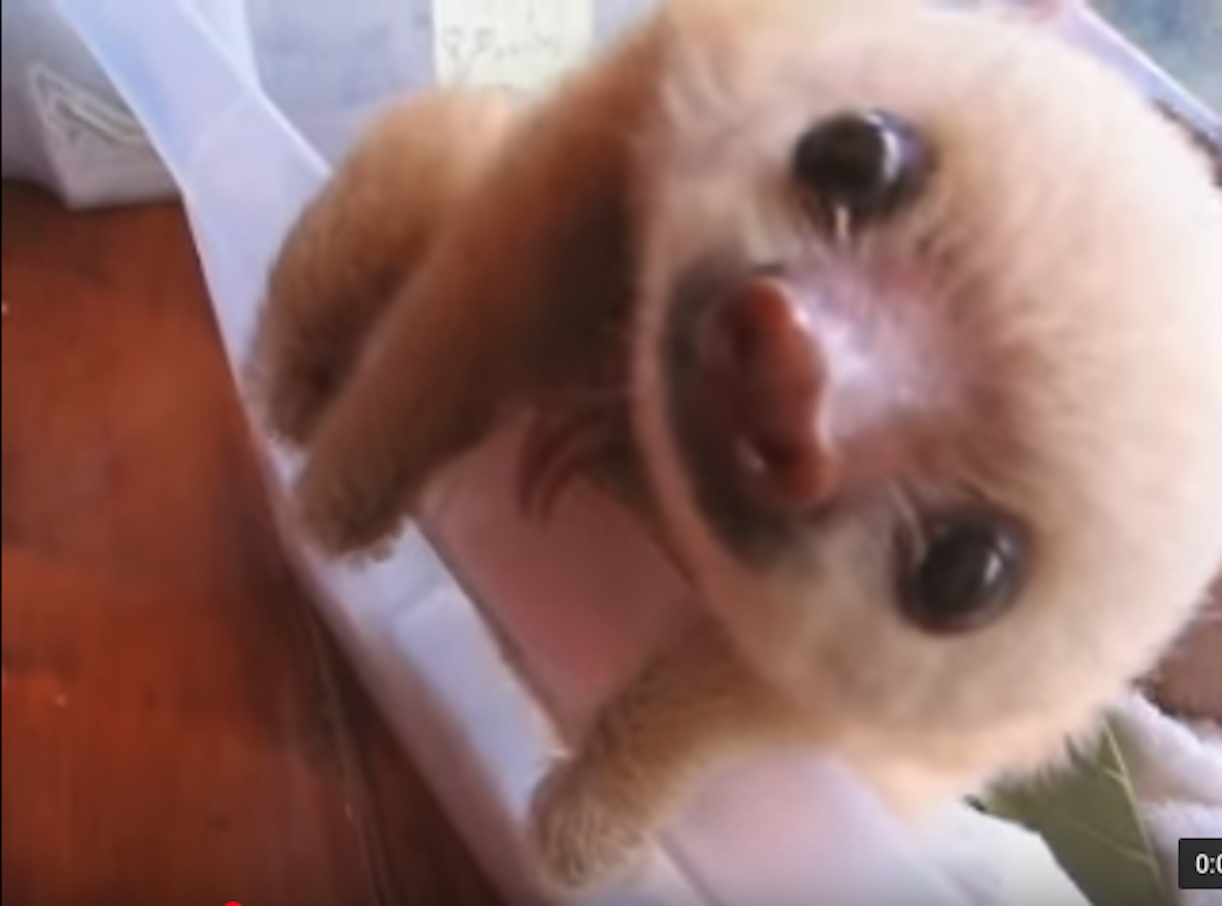} & 
\includegraphics[width=0.55\columnwidth]{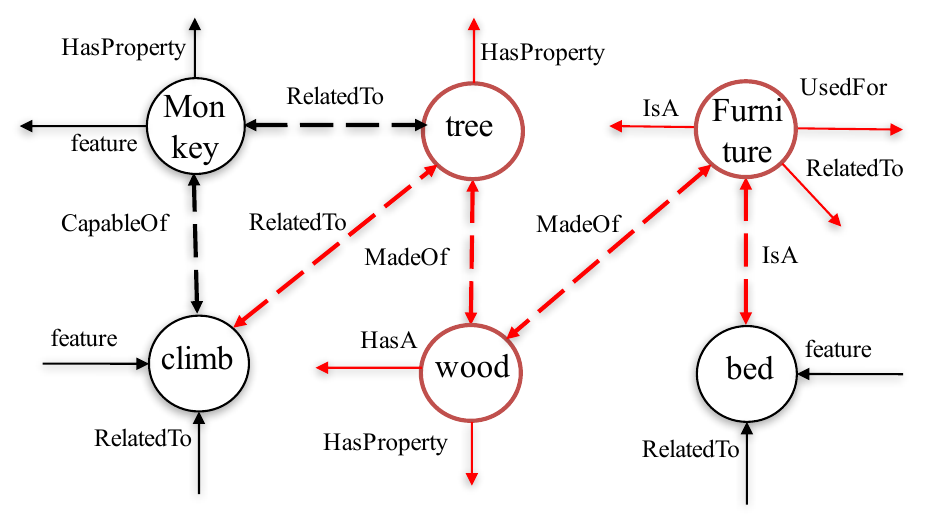} \\

\textit{A person is playing a guitar} 
& \textit{A person is playing a guitar} & 
\textit{A sloth is climbing} & 
\textit{A monkey is climbing}\\
& & \textit{the side of a crib} & \textit{on the bed}

\\
\multicolumn{2}{c}{(a)} & \multicolumn{2}{c}{(b)}\\

\includegraphics[width=0.30\columnwidth]{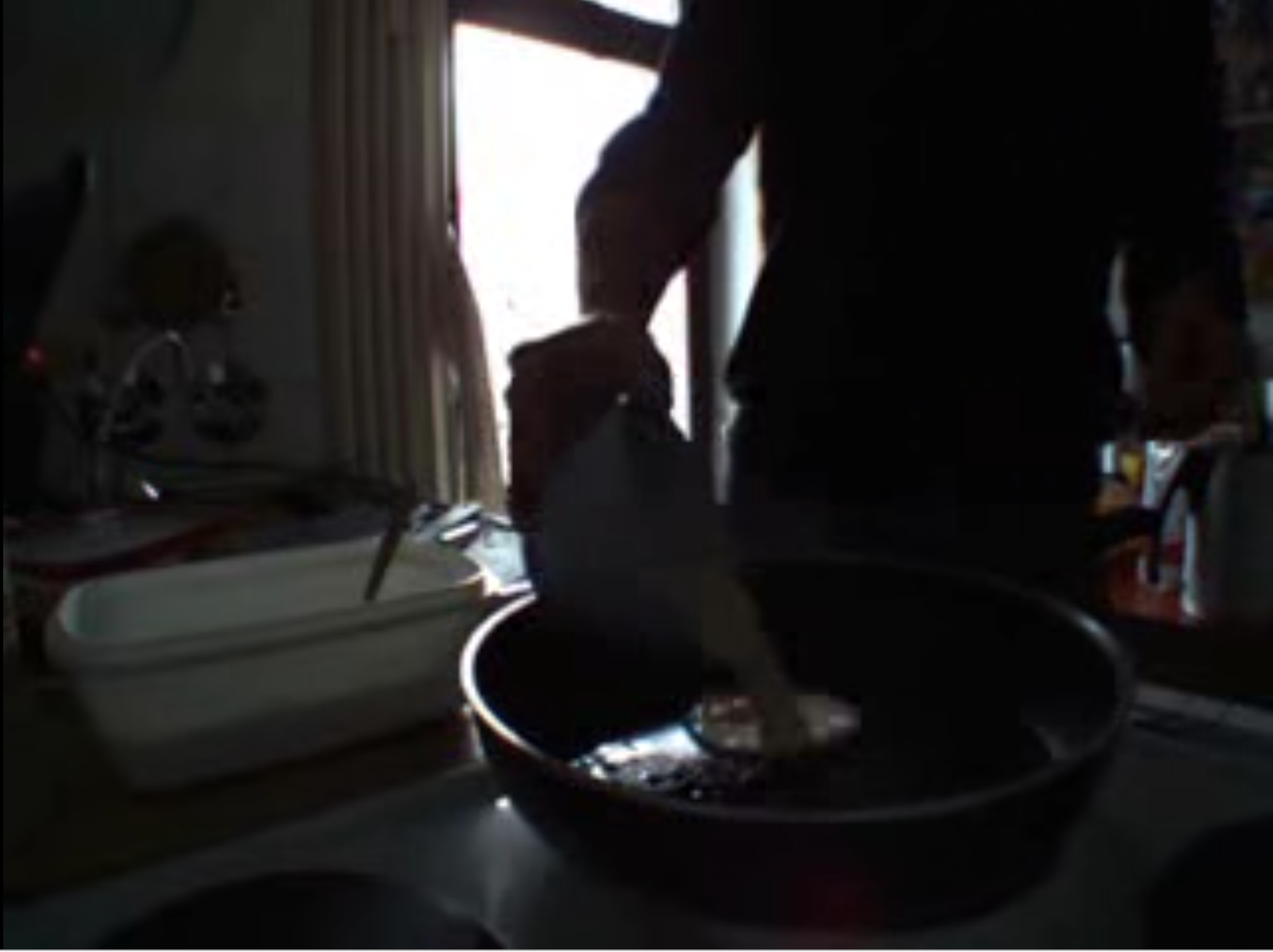} & \includegraphics[width=0.55\columnwidth]{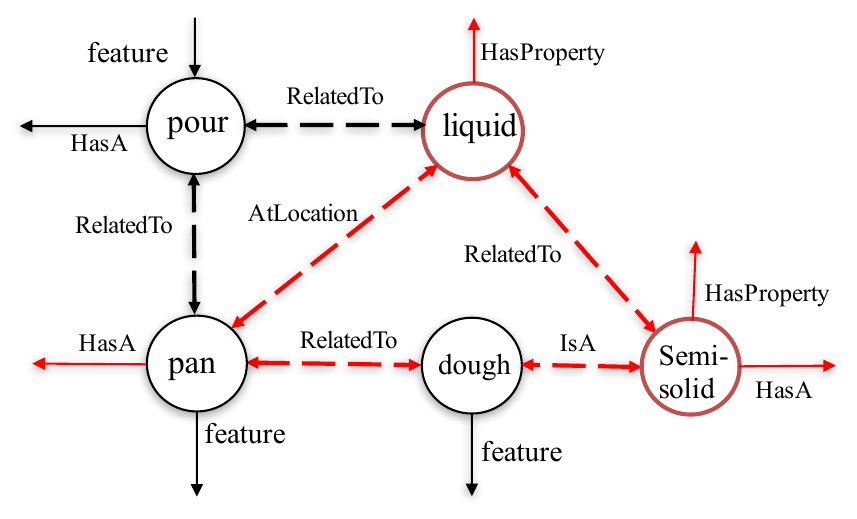} &
\includegraphics[width=0.30\columnwidth]{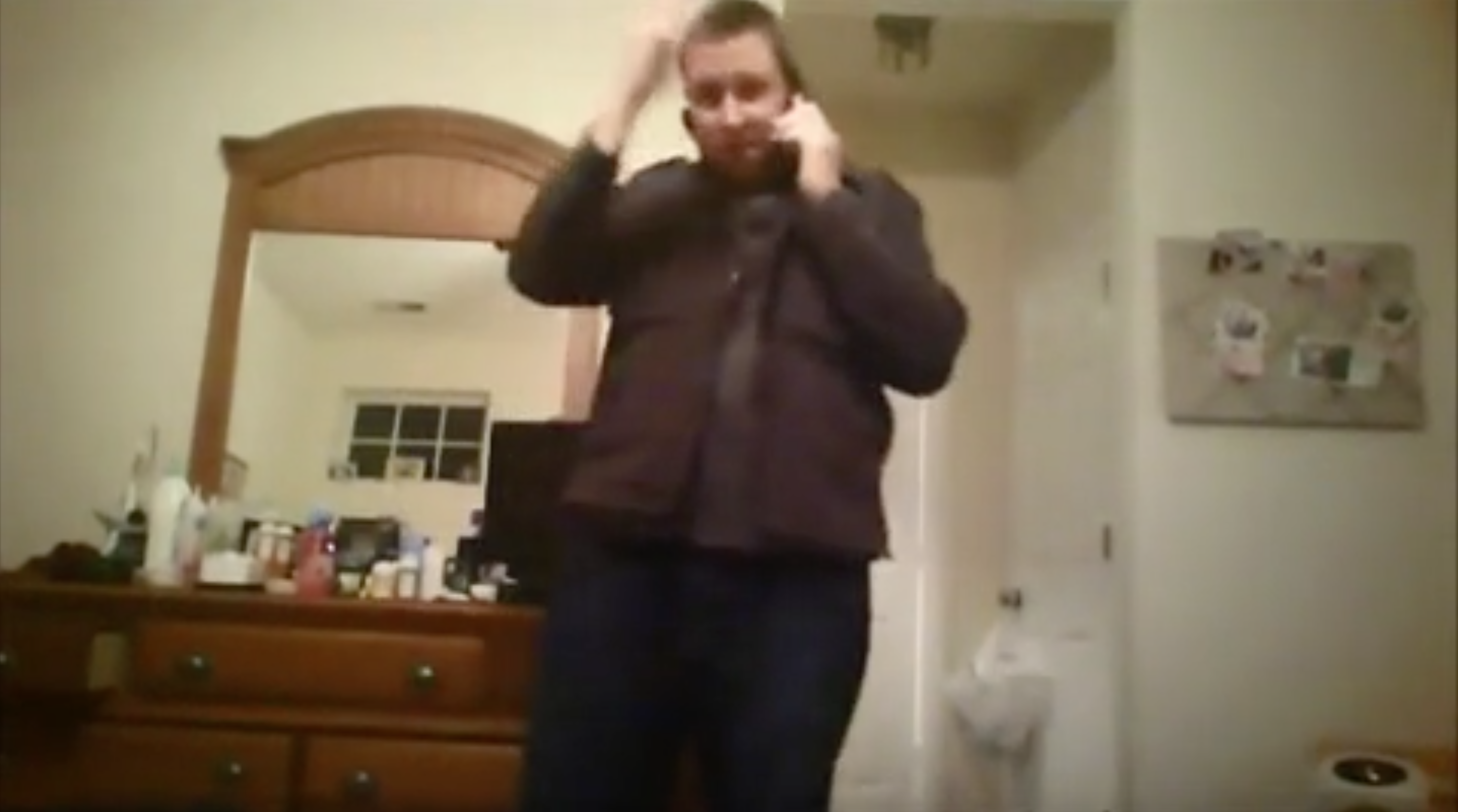} &  
\includegraphics[width=0.55\columnwidth]{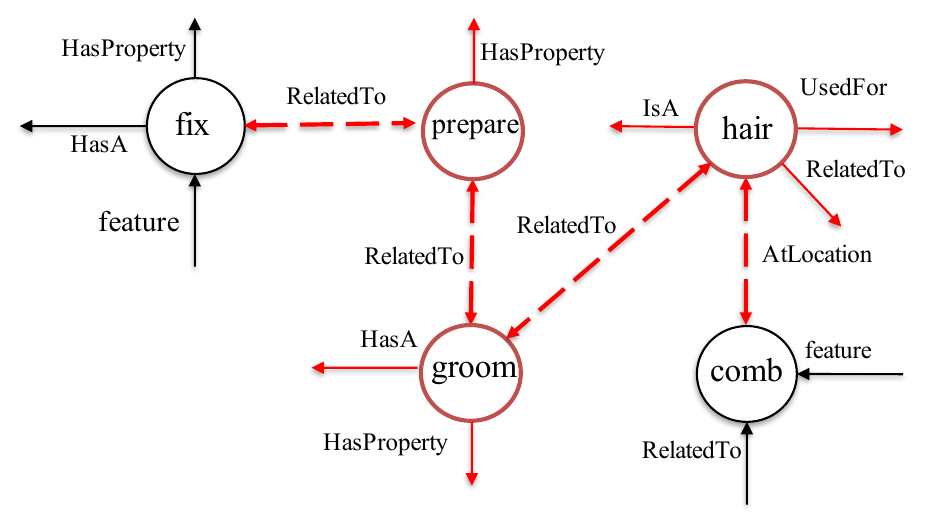} \\

\textit{Pour dough on pan} & \textit{Pour dough on pan} &
\textit{Fix hair} & 
\textit{Fix hair}\\
\multicolumn{2}{c}{(c)} & \multicolumn{2}{c}{(d)} \\

\includegraphics[width=0.20\columnwidth]{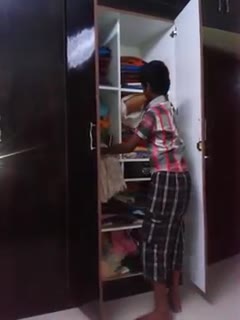} & 
\includegraphics[width=0.55\columnwidth]{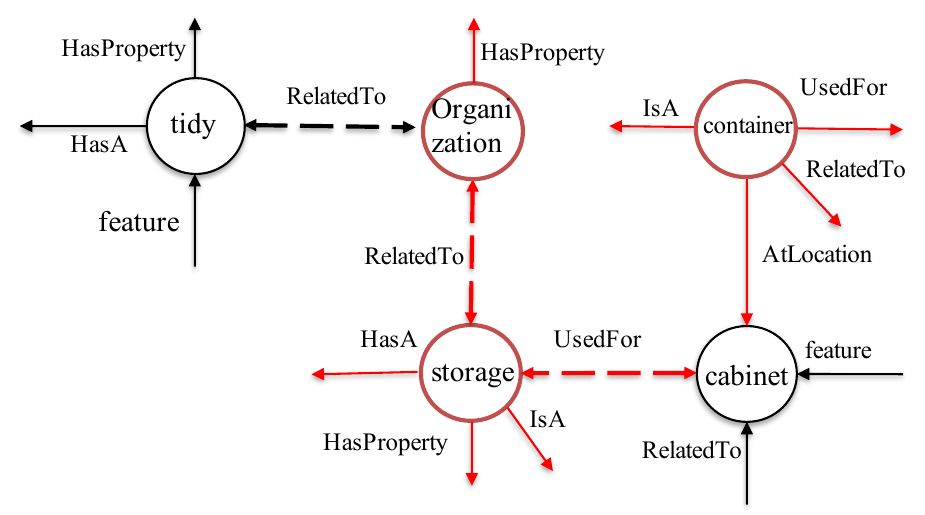} &
\includegraphics[width=0.35\columnwidth]{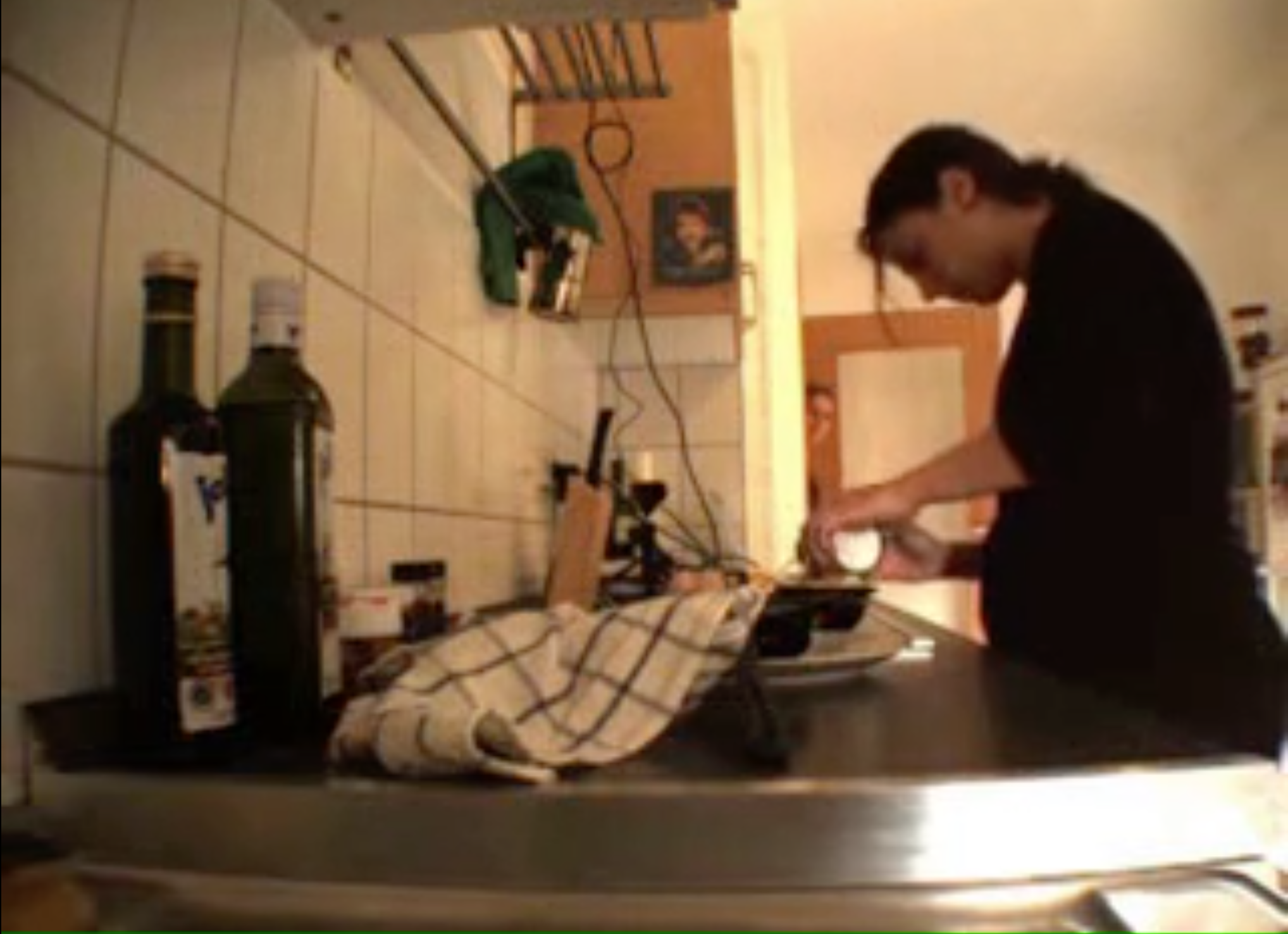} &  
\includegraphics[width=0.55\columnwidth]{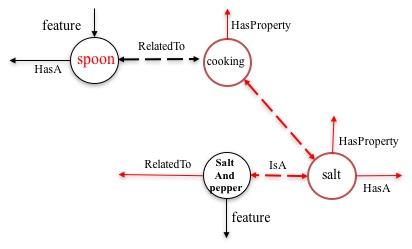} \\
\textit{Tidy cabinet} & 
 \textit{Tidy cabinet} & \textit{Add salt and pepper} & 
\textit{Spoon salt and pepper} \\
\multicolumn{2}{c}{(e)} & \multicolumn{2}{c}{(f)}  \\

\end{tabular}
\caption{Interpretations generated for different input videos from the Breakfast Actions and Charades datasets for the activity recognition task and output interpretations on the MSVD dataset for the video captioning task are shown. Below each interpretation, the automatically generated caption is shown.}
\label{OntologyPerformanceExample}
\end{figure*}
\subsection{Qualitative Analysis}
The use of ConceptNet and contextualization allows the proposed approach to generate semantically rich and coherent interpretations for video activities across domains without having to train explicitly for the target domain. 
As observed from Figure \ref{OntologyPerformanceExample}, the representations produced by our approach are richer with contextualization cues that go beyond what is seen in the image. 
For example, in Figure~\ref{OntologyPerformanceExample}(e), when presented with a video with groundtruth as ``\textit{tidy cabinet}'', our approach was able to relate the presence of the grounded concepts \textit{tidy} and \textit{cabinet} through the ungrounded contextual generators \textit{storage} and \textit{organization}. 
It is also to be noted that for many of the interpretations, the label with the highest confidence score was not the one used in its final (best) interpretation. For example, the confidence scores of the action and object labels \{\textit{stir}, \textit{teabag}\} was higher than the groundtruth combination \{\textit{add}, \textit{teabag}\}. 
However, the contextualization cues allowed for such labeling errors to arrive at the correct interpretation.

It is also interesting to note that the use of an external, common-sense knowledge base in ConceptNet allows us to transcend the stricter vocabulary of the target domain. For example, in Figure~\ref{OntologyPerformanceExample}(f), when presented with an input video with groundtruth ``\textit{Add salt and pepper}'', combined with weaker, handcrafted features, the approach was able to arrive at an interpretation ``\textit{Spoon salt and pepper}''. While the interpretation differed from the groundtruth, it was able to preserve the underlying semantics of the activity without compromising the semantic integrity of the visual scene. 
\subsection{Quantitative Evaluation}
\textbf{Complex Visual Data}: 
For evaluation of the performance on data with complex semantic relationships, we use the \textbf{Charades} dataset \cite{sigurdsson2016hollywood} and report the Mean Average Precision (mAP) score to evaluate the performance of our approach on the recognition task and compare against other comparable approaches. It can be seen from Table \ref{charades_Performance_Table} that the proposed approach has outperformed other state-of-the-art approaches. The Asynchronous Temporal Fields (ATF) approach \cite{sigurdsson2016asynchronous} factors both sequential temporal information and intent. 
It is worth noting that even without the use of intent and temporal sequencing, the proposed approach is able to outperform the state-of-the-art approaches.
\begin{table}[h]
\centering
\begin{tabular}{|c|c|}
\hline
\textbf{Approach}     & \textbf{mAP} \\ \hline
LSTM & 17.80\%              \\ \hline
ATF + trained semantics & 17.30\%              \\ \hline
ATF + trained semantics + trained intent & 17.40\%              \\ \hline
ATF + trained semantics, + trained temporal & 17.40\%              \\ \hline
ATF + trained semantics + intent + temporal & 22.40\%              \\ \hline
PT + ConceptNet & \textbf{29.69}\%              \\ \hline
LSTM + PT + ConceptNet & \textbf{32.56}\%              \\ \hline
\end{tabular}
\caption{Results on Charades dataset. ATF refers to Asynchronous Temporal Fields~\cite{sigurdsson2016asynchronous}. PT + ConceptNet semantics refers to the approach in this paper. Trained semantics indicates use of annotations to capture semantics. 
}
\label{charades_Performance_Table}
\end{table}

\textbf{Complex Semantic Relationships:}
For evaluation of the performance on data with complex semantic relationships, we use the \textbf{Microsoft Video Description Corpus (MSVD)} dataset and report the BLEU score to compare against other comparable approaches. Captions were generated through methods outlined in Section \ref{genCaption}.
Our proposed approach has competitive performance and is very close to the top performing approach as seen from Table \ref{table:MSVD}. 

\begin{table}[h]
\centering
\begin{tabular}{|c|c|c|}
\hline
\textbf{Approach}     & \textbf{BLEU Score} & \textbf{METEOR} \\ \hline
Factor Graph Model \cite{thomason2014integrating}                    & 13.68\%              & 23.9            \\ 
S2VT \cite{venugopalan2014translating}                   & 31.19\%             & 29.2          \\ 
S2VT + COCO \cite{venugopalan2014translating}             & 33.29\%           & -    \\ 
LSTM + Enc-Dec \cite{yao2015describing}             & 41.92\%                &  29.5           \\ 
HRNE \cite{Pan_2016_CVPR}             & 43.6\%              &  32.1          \\
Joint-BiLSTM \cite{bin2016bidirectional} & - & 30.3 \\
aLSTMs (2D) \cite{guo2016attention} & 42.36\% & 29.76\\
aLSTMs (3D) \cite{guo2016attention} & 44.87\% & 30.38\\
PT + ConceptNet & {\bf 42.98}\%       & {\bf 29.6}            \\ \hline
\end{tabular}
\caption{Results on the Microsoft Video Description Corpus (MSVD) dataset. 
}
\label{table:MSVD}
\end{table}

\textbf{Reduced Training Needs:} It is important to note that the training needs of other approaches are significantly higher than ours as they use more complex features that are more descriptive of the concepts within the video.
For example, the HRNE model \cite{Pan_2016_CVPR} makes use of temporal characteristics in the video; the Temporal Attention model \cite{yao2015describing} leverages the frame-level representation from GoogleNet \cite{szegedy2015going} as well as video-level representation using a 3D CNN trained on hand-crafted descriptors along with a dynamic attention model for generating descriptions. 
It is also interesting to note that our approach is also competitive against methods that use strong features (C3D) gained through transfer learning across different domains and datasets \cite{guo2016attention}. 
We, by contrast, just use mean-pooled CNN features across frames to demonstrate the power of using semantic contextualization with prior common-sense knowledge.

\textbf{Weak Features and Uncertainty:} To evaluate the performance of the approach on data with more uncertainty and to demonstrate the use of knowledge, we evaluate on the Breakfast Actions dataset with handcrafted features (HOF and HOG from Section \ref{grounded}). 
The Breakfast Actions dataset introduces several challenges such as varying viewpoints, occlusion and lower visual quality. These complications, when used along with weaker, handcrafted features in HOG and HOF, offer much more uncertainty in the input data and hence provides more confusion among concepts for the underlying machine learning concept recognition models. 
The proposed approach outperforms other state of the art approaches as seen from Table \ref{ba_Performance_Table}. 
It is important to note that our approach is neither trained specifically for the kitchen domain nor on the dataset itself other than for obtaining the starting grounded action and object labels. 
Other methods are restricted by the vocabulary of the training data to build their descriptions. 
For example, the Context Free Grammar and ECTC approaches makes use of temporal information such as transitions between activities to build their final descriptions. 

\begin{table}[h]
\centering
\begin{tabular}{|c|c|}
\hline
\textbf{Approach}     & \textbf{Precision} \\ \hline
HMM \cite{kuehne2014language}                   & 14.90\%              \\ 
CFG + HMM \cite{kuehne2014language}             & 31.8\%               \\ 
RNN + ECTC \cite{huang2016connectionist} & 35.6\% \\
RNN + ECTC (Cosine) \cite{huang2016connectionist} & 36.7\% \\
HCF + PT+weights & 33.40\%              \\ 
HCF + PT+training \cite{desouza2016spatially}  & 38.60\%              \\ 
HCF + PT + ConceptNet & \textbf{42.98}\%              \\ \hline
\end{tabular}
\caption{\textit{Handling weaker features:} Results on Breakfast Action dataset.
HCF refers to the use of handcrafted features from Section~\ref{grounded}}
\label{ba_Performance_Table}
\end{table}

{\bf Immunity to Unbalance in Training Data:} Not all labels are equally represented in most training data. This is particularly acute as number of labels increase. To demonstrate that our method is immune to this effect, we partitioned the activity class labels from the Breakfast Actions dataset into 4 different categories their frequency in training data. Table~\ref{table:unbalance} shows the performance for our approach as compared to prior pattern theory approaches, PT+training and PT+weights. As expected, performance of PT+training that rely on annotations~\cite{desouza2016spatially} increase with increase in training data, whereas, our approach is stable. It is interesting to note that this non-uniform class distribution also affects the performance of the machine learning models that provide the plausible concept labels. However, the use of ConceptNet and its rich semantic structure allows us to overcome this challenge and maintain the performance across varying amounts of training data. 
\begin{table}[h]
\centering
\begin{tabular}{|c|c c c c|}
\hline
\textbf{Approach} & \multicolumn{4}{c|}{\textbf{Num. of  samples}}\\ \cline{2-5}
 & $ \le 10$ & $10-20$ & $20-40$ & $\ge 40$   \\  \hline
PT+training \cite{desouza2016spatially} & 11.8\%	& 17.4\% & 22.9\% & 35.5\% \\ 
PT+weights &	25.4\% & 36.6\% & 37.1\% & 34.26\% \\ 
PT+ConceptNet &	34.8\% & 40.1\% & 38.2\% & 38.4\% \\ 
\hline
\end{tabular}
\caption{\textit{Impact of data imbalance:} Comparison of pattern theory approaches with different semantic knowledge sources for different activity categories that differ in number of training samples.}
\label{table:unbalance}
\end{table}

\vspace{-0.5cm}
\section{Conclusion}
Contextualization cues from commonsense knowledge, such as ConceptNet, can be used to generate rich semantic interpretations of video, beyond simple semantic relationships. 
This also helps us break the ever-increasing demands on annotation quality and quantity of training data. 
There is no training required beyond that needed for the starting object and action labels. 
We demonstrate how pattern theory can be used to capture the semantics in ConceptNet and infer rich interpretations that can be the basis for generating of sentences or even visual question and answers. 
The inference process allows for multiple concept labels for each video event to overcome errors in classification. 
Extensive experiments demonstrate the applicability of the approach to different domains and its highly competitive performance. 
\section{Acknowledgement}
This research was supported in part by NSF grant CNS-1513126. 

This work includes data from ConceptNet 5, which was compiled by the Commonsense Computing Initiative. ConceptNet 5 is freely available under the Creative Commons Attribution-ShareAlike license (CC BY SA 4.0) from http://conceptnet.io. The included data was created by contributors to Commonsense Computing projects, contributors to Wikimedia projects, Games with a Purpose, Princeton University's WordNet, DBPedia, OpenCyc, and Umbel. 

{\small
\bibliographystyle{ieee}
\bibliography{egbib}
}

\end{document}